# Semi-Supervised Classification for oil reservoir

Yanan Li
*School of Economics and Management*
*China University of Geosciences*
Wuhan 430074, China
liyanan1990@hotmail.com

Haixiang Guo
*School of Economics and Management*
*China University of Geosciences*
Wuhan 430074, China
faterdumk0732@sina.com

Andrew P Paplinski
*Faculty of Information Technology*
*Monash University*
Melbourne 3800, AU
andrew.paplinski@monash.edu

*Abstract* — This paper addresses the general problem of accurate identification of oil reservoirs. Recent improvements in well or borehole logging technology have resulted in an explosive amount of data available for processing. The traditional methods of analysis of the logs characteristics by experts require significant amount of time and money and is no longer practicable. In this paper, we use the semi-supervised learning to solve the problem of ever-increasing amount of unlabelled data available for interpretation. The experts are needed to label only a small amount of the log data. The neural network classifier is first trained with the initial labelled data. Next, batches of unlabelled data are being classified and the samples with the very high class probabilities are being used in the next training session, bootstrapping the classifier. The process of training, classifying, enhancing the labelled data is repeated iteratively until the stopping criteria are met, that is, no more high probability samples are found. We make an empirical study on the well data from Jianghan oil field and test the performance of the neural network semi-supervised classifier. We compare this method with other classifiers. The comparison results show that our neural network semi-supervised classifier is superior to other classification methods.

I. INTRODUCTION

As the demand for oil grows steadily, many countries including China increases their effort to identify new oil reservoirs as an answer to significant oil shortages [1]. Recent improvements in well logging technology result in significant increase of logging data that must be automatically processed and classified. The traditional methods used to classify the oil reservoirs, are based on the analysis of the logs' characteristics by experienced experts who assign class labels to samples from the datasets. Such a method is prone to simplifications and errors, has an obvious time limitations and generates difficult to satisfy demand for highly trained experts.

Traditional classification method in the Oil Industry is known as the History Matching (HM) problem [2]. A model of accurate forecast for the future behaviour of the reservoir can improve reservoir decision making and save huge investments in the oil industry. Guo et al. [3] presented a method of feature selection to construct a model to match the log data and forecast new log data. Aranha et al. [4] used tuned evolutionary algorithms and adaptive differential evolution to optimize the models of oil reservoirs. These approaches could improve the classification results compared to the standard methods presented in the HM literature. Hutahaean et al. [5] presents the vector-guided evolutionary algorithm (RVEA) [6] that can be used to address the HM problem, and demonstrates the superiority of RVEA to the state of the art history matching algorithms. Lino et al. [7] used rapid multi-phase simulation with the Fast Marching Method (FMM) for solving the HM problem, and validated the accuracy and computational efficiency of FMM. Zhang et al. [8] proposed history matching with dimensionality reduction based on Multi-objective Evolutionary Algorithm Based on Decomposition, aiming at handling the pre-image problem and improving model performance. Most of these methods used to construct a classification model need a dataset with labelled samples. However, the amount of data available from the oil reservoirs is very large and rapidly increasing, and most data from well logging are unlabelled.

To alleviate the problem of classification of mainly unlabelled data, the semi-supervised learning (SSL) can be used. The method originates from the self-training classification method [9] and is recently actively researched, see for example [10], [11], [12]. The general idea of semi-supervised method is to train the classifier first with a small set of available labelled data and then iteratively retrain the classifier with the big set of unlabelled data.

In this paper, we use a raw well-logging data from the Jianghan oilfield to test the performance of a semi-supervised neural network classifier. Firstly, the classifier is trained using the dataset containing samples with labels. Next, this initially trained model is used to classify datasets without labels. As a result, each unlabelled sample is allocated probabilities of belonging to specific classes. The samples with very high class probabilities are assigned the class labels and are added into the labelled dataset. Then, the neural network classifier is retrained iteratively until the termination condition is satisfied. In this

This work is partially supported by NSF Grant #71103163, #71103164, #71573237

way, we are able to classify and label samples from the unlabelled datasets.

The rest of this paper is organized as follows. In section 2, we explain our dataset and describe details of the semi-supervised neural network classifier. In section 3, we describe our experiments and analyse the results. Finally, in section 4, we summarize the paper and discuss future work.

## II. USE EXPERIMENTAL SETUP

### A. Data sets

In this paper, we analyse the dataset *oilsk81* that originates from well-logging data from the Jianghan oilfield. The dataset consists of 12,154 samples with 17 attributes. The attributes, or features, depend on the specific well logging technology. Our data set includes the following attributes: depth of the well (D), eresistivity (R2.5), eresistivity (R0.5), deep lateral resistivity (LLD), shallow lateral resistivity (LLS), compensate neutron logging (CNL), acoustic time (AC), Caliper (CAL), Caliper (CAL1), Caliper (CAL2), Caliper (CALC), hole deviation angle (DEVI), hole azimuth angle (AZIM), clay (CL), gamma ray (GR), shale index (SH), well temperature (TEMP).

The problem is to identify the characteristics of each geological layer described by the attributes and the ultimate goal is to find the oil layer (O) and the inferior layer (I). In addition to the above layers, it is also possible to identify the dry layer (D) and the water layer (W). Hence, all together, the samples will be classified into the four classes labelled {D, W, I, O}.

The initial step involves human experts that can identify a small subset of the samples that they can classify, or label with high accuracy. In the case of the *oilsk81* dataset, out of the total 12,154, the experts labelled only 496 samples.

Table I describes the characteristics of the dataset, both the subset containing the labelled samples and samples without label. The rows illustrate samples from the above two subsets, whereas the columns represent the total number of samples in each subset (#E), and the number of the attributes (#A) and classes (#C). The four following columns give the number of samples in each class: class D (#D), class W (#W), class I (#I) and class O (#O). Looking at the distribution of samples in classes, we note that it is rather unbalanced. The number of samples in the class W is significantly smaller than in other classes. It is worth emphasizing again that the number of samples without labels is much greater than the number of samples with labels.

TABLE I. CHARACTERIZATION OF THE *OILSK81* DATASET

| Dataset | #E | #A | #C | #D | #W | #I | #O |
|---|---|---|---|---|---|---|---|
| Subset with labels | 496 | 17 | 4 | 162 | 50 | 107 | 177 |
| Subset without labels | 12154 | 17 | 4 | ? | ? | ? | ? |

The question marks in Table I will be filled in using the classifiers discussed in the subsequent sections.

### B. General description of the semi-supervised neural network classifier

Our semi-supervised classifier is a simple Artificial Neural Networks (ANN) [13] consisting of two neuronal layers [14, 15]. The hidden layer uses the ReLU as the activation function. The output layer uses the softmax function to calculate for each sample the probabilities of belonging to each class.

In the pre-processing step we first normalize the dataset and then split into two subsets, one with the labelled samples and the other one with the unlabelled samples as in Table I.

The top-level description of our method consists of two phases. In the first, or supervised phase, we train our neural network classifier with the labelled data. The obtained results will be referred to as a NN-supervised classifier. We start with the initialization of the hidden and output weight matrices. Then, for each sample:

1. In the feedforward phase, we calculate the outputs from the hidden and the output layers,
2. In the backpropagation phase, we calculate the delta errors and the gradients of the loss/objective function layer-by-layer, and update the weights matrices accordingly.
3. Check the termination condition, which is typically determined by the value of the loss function.

Once the training is completed, we calculate the accuracy of training as the ratio of correctly classified samples to the total number of samples. We used a 5-fold validation method as described below.

In the second, or semi-supervised phase [9], we start with classification of the unlabelled samples by the NN-supervised classifier. There are two problems that need to be addressed in this phase. The first one is the unbalanced distribution of samples into classes. For the labelled data the distribution of samples is shown in Table 1. The second problem is that many unlabelled samples are only weakly classified, that is, no prominent class probability can be identified. The general idea of the semi-supervised learning is that the strongly classified unlabelled data samples are labelled according to the highest class probability and attached to the labelled data set. We have an opportunity to balance the new training set by repeating the number of samples from the under-represented classes. The NN-semi-supervised classification involves the following steps:

1. Train the neural network classifier for the current set of the labelled data,
2. classify the remaining unlabelled data finding the maximum class probabilities corresponding to every sample,
3. Select strongly classified samples according to selection criteria, and attached them to the labelled subset.

4. Repeat the training and selection until the termination criteria are met.

The selection criteria in Step 3 that determine the strong classification, are crucial to the performance of the semi-supervised learning. The specific criteria are adjusted according to the results but are based on the values of the class probabilities. Termination, ultimately, takes place when all samples are labelled, or when only unlabelled samples left have weak class probabilities.

## III. EXPERIMENTS AND ANALYSIS

In order to test the performance of our neural network based classifiers we conduct two sets of testing experiments. Firstly, the NN-supervised classifier is compared with a number of other classifiers, namely, the Discriminant analysis classifier [16], KNN classifier [17], NaiveBayes classifier [18], Ensembles classifiers [19], and SVM classifier [20] using the expert labelled data. In the second test, the NN-semi-supervised classifier is compared with the above six classifiers using the unlabelled data.

### A. Parameters setup

In order to obtain statistically significant results we use the 5-fold cross-validation method and repeat each learning run 25 times for each subsets of the labelled data – total of 125 runs. Each learning run involves 50,000 learning epochs to guarantee the small final value of the loss/error function. Each learning run takes 3.3 minutes.

The learning rates have been experimentally selected using the sensitivity analysis. The learning rate for the hidden layer is set as 50/N (where the N is the number of data samples), the learning rate for the output layer as 35/N. It seems that regularization does not influence the learning results. From the structural point of view, we have 50 neurons in the hidden layer and 4 neurons in the output softmax layer [15]. In the KNN classifier, we set the number of nearest neighbours to be equal to 1.

### B. Evaluation criterion

For every learning run, we calculate the resulting classification accuracy. The accuracy is calculated as the ratio of the number of samples which are correctly classified to the total number of samples in dataset used for testing. To calculate the effective performance of the given classifier, for all 125 learning runs we calculate the mean accuracy, the max accuracy, and the min accuracy

### C. Selection criteria for the semi-supervised classifier

After each training session as described above, the neural network classifier is presented with the unlabelled data samples. As describe before, at this stage we select the samples that are considered to be correctly classified according to a selection criterion. Selected samples are attached to the pool of labelled data for subsequent training. This procedure replaces the human expert typically used to label the raw dataset and is the essence of the semi-supervised learning. As the selection criterion, we use the classification probabilities calculated by the neural network. For this purpose, we divide the range of probabilities into eight groups labelled: V-poor, Poor, NT-poor, NT-good, Good, V-good, VV-good, and VVV-good, respectively. See Table V for more details.

### D. Results of training the classifier with the labelled subset

The accuracies of training the selected classifiers with the labelled data are shown in Table II. As explained above we use the 5-fold cross-validation method and repeat each learning run 25 times. Table II demonstrate the fact that accuracy of the supervised neural network classifier is the highest for the test dataset, the KNN classifier delivers the second best accuracy, as marked in bold in Table II. Although the KNN classifier performs a tiny little bit better for the training data, by increasing the number of parameters in the NN-supervised classifier, it could also achieve the perfect score for the training data. All other classifiers deliver significantly worse training and testing accuracies.

TABLE II.  ACCURACIES FOR THE TRAINING AND TESTING OF THE LABELLED SUBSET OF THE *OILSK81* DATASET

| Algorithm | Mean accuracy | Max accuracy | Min accuracy |
|---|---|---|---|
| Discriminant analysis | 0.8916/0.8723 | 0.9395/0.9697 | 0.8615/0.7900 |
| KNN | **1/0.9680** | **1/1** | **1/0.9200** |
| NaiveBayes | 0.7153/0.6978 | 0.7481/0.7980 | 0.6902/0.6061 |
| Ensembles | 0.7686/0.7417 | 0.8086/0.8384 | 0.7128/0.6162 |
| SVM | 0.9348/0.9061 | 0.9520/0.9596 | 0.9194/0.8485 |
| NN-supervised classifier | 0.9999/**0.9703** | **1/1** | 0.9949/**0.9292** |

Note: The value before '/' is the accuracy for the training dataset, and the value after '/' is the accuracy the test dataset.

In Table III, we present another view of classification accuracy for the labelled dataset by giving the numbers of samples classified into each group. The number corresponds to the minimum accuracy of Table II. The last row gives the ground truth. It is clearly visible that the Discriminant analysis classifier and the Ensembles classifier perform worse on class D and class I, and the NaiveBayes classifier performs worse on class D, class I, and class O. The SVM classifier delivers some intermediate results of accuracy.

TABLE III.  RESULT OF CLASSIFICATION LABELS ON DATASET WITH LABEL

| Classifier | #D | #W | #I | #O |
|---|---|---|---|---|
| Discriminant analysis | 117 | 57 | 152 | 170 |
| KNN | 161 | 50 | 107 | 178 |
| NaiveBayes | 140 | 48 | 57 | 152 |
| Ensembles | 184 | 55 | 72 | 185 |
| SVM | 166 | 51 | 117 | 162 |
| NN-supervised | 160 | 51 | 107 | 178 |
| Expert Labelled Dataset | 162 | 50 | 107 | 177 |

In Table IV we show the numbers of misclassified samples for each classifier trained and tested for the labelled dataset. For each class we show where the misclassified samples go. All but NaiveBayes classifier correctly identify the W (water) class samples. The NaiveBayes puts one W sample in the O (oil) class. The Discriminant analysis classifier tends to misclassify the I-class samples. The KNN classifier misclassifies the small number of class I and O samples. The weaknesses of other classifiers can be easily identified from Table IV. Most importantly, the NN-supervised classifier identifies correctly the oil (O) samples.

TABLE IV. NUMBERS OF MISS-CLASSIFIED SAMPLES FOR DIFFERENT CLASSIFIERS FOR THE LABELLED DATASET

| Classifier | class | D | W | I | O | Total of misclassified samples |
|---|---|---|---|---|---|---|
| Discriminant analysis classifier | D | 0 | 7 | 37 | 6 | 50 |
|  | W | 0 | 0 | 0 | 0 | 0 |
|  | I | 1 | 0 | 0 | 0 | 1 |
|  | O | 4 | 0 | 0 | 0 | 4 |
| KNN | D | 0 | 0 | 1 | 3 | 4 |
|  | W | 0 | 0 | 0 | 0 | 0 |
|  | I | 0 | 0 | 0 | 1 | 1 |
|  | O | 3 | 0 | 0 | 0 | 3 |
| NaiveBayes | D | 0 | 7 | 52 | 21 | 80 |
|  | W | 0 | 0 | 0 | 1 | 1 |
|  | I | 5 | 0 | 0 | 2 | 7 |
|  | O | 29 | 0 | 26 | 0 | 55 |
| Ensembles | D | 0 | 5 | 34 | 21 | 60 |
|  | W | 0 | 0 | 0 | 0 | 0 |
|  | I | 69 | 0 | 0 | 0 | 69 |
|  | O | 13 | 0 | 0 | 0 | 13 |
| SVM | D | 0 | 1 | 11 | 3 | 15 |
|  | W | 0 | 0 | 0 | 0 | 0 |
|  | I | 8 | 0 | 0 | 0 | 8 |
|  | O | 11 | 0 | 7 | 0 | 18 |
| NN-supervised classifier | D | 0 | 1 | 0 | 3 | 7 |
|  | W | 0 | 0 | 0 | 0 | 0 |
|  | I | 2 | 0 | 0 | 1 | 3 |
|  | O | 0 | 0 | 0 | 0 | 0 |

Note: Every row gives the numbers of misclassified samples that are incorrectly allocated to the class specified by the respective column. For example, for the Discriminant Analysis Classifier, the first row indicates that 7 D-class samples have been incorrectly allocated to class W, 37 − to class I and 6 − to class I. In total, 50 D-class samples have been misclassified number.

### E. Probability analysis of applying the NN-supervised classifier for the labelled and unlabelled datasets.

In sec. IIID, we trained a NN-supervised classifier model with the subset with labelled samples. Table V shows the class probabilities for every sample in the labelled data subset, and the distribution of probabilities between the eight probability classes as described in sec. IIIC. Note that the higher the class probabilities, the stronger the reliability of classification. From Table V and Fig.1, it can be seen that for each class, the majority of samples are classified with probabilities greater than 0.6. In total, 80% of samples in all classes are classified with probabilities greater than 0.9. Most importantly, 91% of class O (oil) samples are also classified with probabilities above 0.9. Overall, we can conclude that the accuracy of NN-supervised classifier for the labelled samples is very high. This gives a good prospect of reliable classification of the unlabelled data.

TABLE V. THE CLASS PROBABILITIES OF THE SAMPLES FROM THE **LABELLED** DATASET.

| Performance | Range of probability | Number of samples all in classes | #D | #W | #I | #O |
|---|---|---|---|---|---|---|
| VVV-good | (0.95,1.0] | 343 | 108 | 42 | 50 | 143 |
| VV-good | (0.90,0.95] | 52 | 11 | 3 | 22 | 16 |
| V-good | (0.80,0.90] | 51 | 25 | 4 | 16 | 6 |
| Good | (0.70,0.80] | 24 | 7 | 1 | 11 | 5 |
| NT-good | (0.60,0.70] | 15 | 6 | 0 | 6 | 3 |
| NT-poor | (0.50,0.60] | 10 | 3 | 0 | 6 | 1 |
| Poor | (0.40,0.50] | 1 | 0 | 0 | 0 | 1 |
| V-poor | [0.0,0.40] | 0 | 0 | 0 | 0 | 0 |
| sum |  | 496 | 160 | 50 | 111 | 175 |

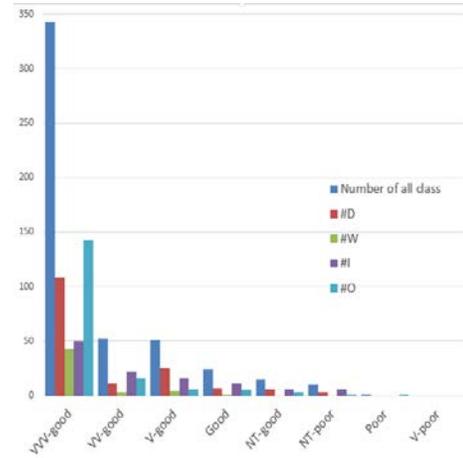

Fig. 1. Distribution of the class probabilities of the samples from the **labelled** dataset.

In Table VI and Fig.2, we illustrate the probability distributions of unlabelled samples into four classes. The percentage of samples in each group that have been classified the probabilities above 0.8 is 88%, 28%, 8% and 37%, respectively. This fact demonstrates that proper classification of the unlabelled samples is not a trivial task. However, most of the samples are classified with probabilities greater than 0.5. Conversely, on the other side of the spectrum, only 1% of samples are classified with probability smaller than 0.4. In each range of probabilities, the class D has the largest number of samples, especially when the probabilities are larger than 0.5. Distribution of samples belonging to class O peaks for NT-good probabilities, that is, in the range of (0.60, 0.70].

It is important to note that the number of samples which are labelled as class D is much bigger than the number of samples from all other classes, especially class W and class I. A probable reason for this imbalance is that in the original labelled dataset the class D samples are most numerous. We address this information imbalance, in such a way that for the subsequent semi-supervised training steps, we select only samples belonging to the class W and class I classified with probabilities greater than 0.7.

TABLE VI. THE CLASS PROBABILITIES OF THE SAMPLES FROM THE **UNLABELLED** DATASET.

| Performance | Range of probability | Number of samples all in classes | #D | #W | #I | #O |
|---|---|---|---|---|---|---|
| VVV-good | (0.95,1.0] | 8122 | 7938 | 13 | 0 | 171 |
| VV-good | (0.90,0.95] | 852 | 739 | 15 | 0 | 98 |
| V-good | (0.80,0.90] | 910 | 712 | 11 | 11 | 176 |
| Good | (0.70,0.80] | 704 | 465 | 21 | 28 | 190 |
| NT-good | (0.60,0.70] | 639 | 369 | 13 | 29 | 229 |
| NT-poor | (0.50,0.60] | 648 | 336 | 33 | 38 | 241 |
| Poor | (0.40,0.50] | 242 | 110 | 27 | 19 | 86 |
| V-poor | [0.0,0.40] | 37 | 12 | 6 | 3 | 16 |
| sum | | 12154 | 10681 | 138 | 128 | 1207 |

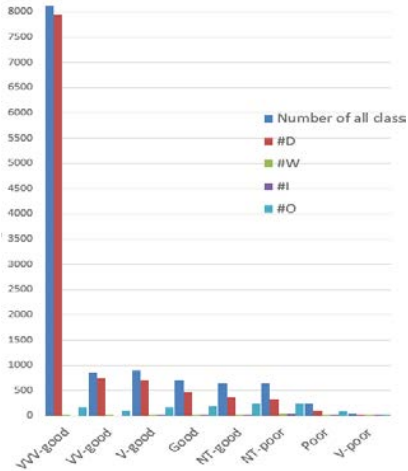

Fig. 2. Distribution of the class probabilities of the samples from the **unlabelled** dataset.

In Table VII we demonstrate the first step of the process of labelling samples during the semi-supervised learning. Comparing with Table I we note that the number of labelled samples increased by 99, but importantly, the newly labelled samples were allocated predominantly to the classes W and I. As a result, the new-labelled dataset in better balanced.

TABLE VII. DISTRIBUTION OF THE LABELLED SAMPLES AFTER THE FIRST STEP OF THE SEMI-SUPERVISED LEARNING

| Dataset | #E | #V | #C | #D | #W | #I | #O |
|---|---|---|---|---|---|---|---|
| Subset with label | 595 | 17 | 4 | 162 | 110 | 146 | 177 |
| Subset without label | 12055 | 17 | 4 | ? | ? | ? | ? |

*F. Dynamic of the semi-supervised classification*

As described in sec. IIB the semi-supervised learning is based on selection of strongly classified samples and attaching them to the current labelled set that will be used in the next supervised learning step referred subsequently to as the updating step. In Fig. 3 we present the dynamic of the semi-supervised classification plotting the numbers of samples in the current labelled set for the updating steps. It can be seen that the number of samples in the labelled dataset increases with the updating steps, while the number of still unlabelled samples is decreasing accordingly. The curve follows the s-shape, having the small increase of the labelled data at the beginning of the process and saturation at the end.

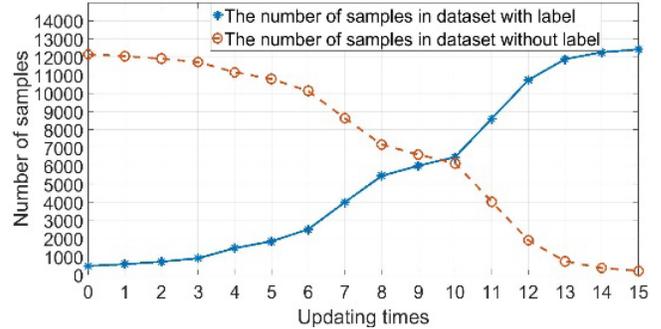

Fig. 3. Change of the number of labelled and unlabelled data samples during semi-supervised larning

In Fig. 4 we show the accuracy after each updating step. In the first part of the semi-supervised learning, the accuracy of classification decreases only slightly. In the second part, when seemingly the strongly classified samples are exhausted, the process of decreasing the accuracy accelerates by the increase number of weekly-classified samples.

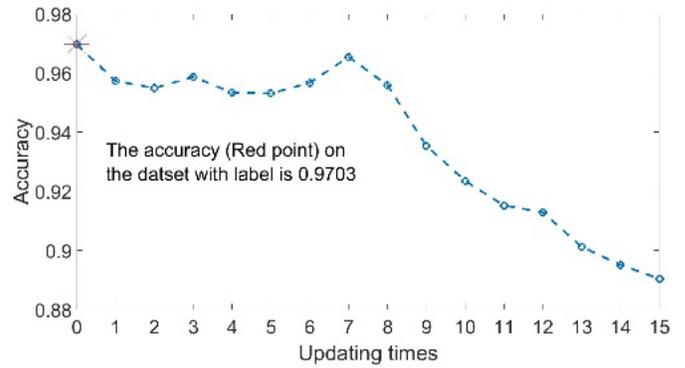

Fig. 4. Change of accuracies by updating times

*G. Comparison of classification results for unlabelled dataset by different classifier. The balance factor.*

In Table VIII, we can see that the number of samples allocated to the class D, class W, class I, and class O differ significantly. It can be seen that, in the Discriminant analysis, NaiveBayes, and SVM classifiers, the number of samples in class I is equal to 0. Therefore, the **balance factor** measured as the ratio of the smallest number of samples in a class to the highest, is equal to 0. In the KNN classifier, the class I has the smallest number of samples, namely, 102. The class D, is the most numerous containing 9036 samples. Hence the balance factor for the KNN classifier is equal to 102/9036 = 1.1%. Similarly, the balance factors to the Ensembles classifier is 710/8740 = 8.1%, and for the NN-supervised classifier

128/10681 = 1.2%. Finally, the NN-semi-supervised classifier has the balance factor 527/4352 = 12.1%. In conclusion, judging by the balance factor, the NN-semi-supervised classifier is superior to five others.

TABLE VIII. COMPARISON OF CLASSIFICATION RESULTS OF THE UNLABELLED DATA FOR DIFFERENT CLASSIFIER

| Classifier | Number of samples in class D | Number of samples in class W | Number of samples in class I | Number of samples in class O | Same number |
|---|---|---|---|---|---|
| Discriminant analysis | 0 | 11998 | 0 | 156 | 4035 |
| KNN | 9036 | 255 | 102 | 2761 | 3576 |
| NaiveBayes | 8053 | 0 | 0 | 4101 | 3377 |
| Ensembles | 710 | 1582 | 1122 | 8740 | 6214 |
| SVM | 8863 | 234 | 0 | 3057 | 3716 |
| NN-supervised | 10681 | 138 | 128 | 1207 | 3552 |
| NN-semi-supervised | 3108 | 4167 | 527 | 4352 | **12154** |

Note: 'Same number' means the number of samples whose labels coming from the corresponding classifier are same as the labels coming from NN-semi-supervised method

## IV. CONCLUSION

Facing ever-increasing amount of log data from the oil well logging, the traditional methods used to identify the oil reservoir that rely on the experts becomes non-viable. In this paper, we use the semi-supervised learning to allocate unlabelled samples to four classes, the O (oil) class in particular. In this method, the experts are needed to label only a small amount of log data. The neural network classifier is first trained with the initially labelled data. Next batches of unlabelled data are being classified and the samples classified with the very high probabilities are being used in the next training session, bootstrapping the classifier. The process of training, classifying, enhancing the labelled data is repeated iteratively until the stopping criteria are met, that is, no more high probability samples are found. We make an empirical study on the well data from Jianghan oil field and test the performance of the neural network semi-supervised classifier. We compare this method with other classifiers and find that our neural network semi-supervised classifier is superior to other classification methods.

Future research will include study of deep neural architecture and evaluation its applicability in classification of data from oil wells.

We hope that experts positively evaluate our improved results of classification the unlabeled well-log data.

## ACKNOWLEDGMENT

This work is supported by the National Natural Science Foundation of China No.71103163, 71103164, 71573237 by Program for New Century Excellent Talents in University, No. NCET-13-1012, by Research Foundation of Humanities and Social Sciences of Ministry of Education of China No.10YJC790071, by the Fundamental Research Founds for National University, China University of Geosciences (Wuhan) No.CUG120111, CUG110411, G2012002A, CUG140604, by China Postdoctoral Science Foundation Grant No. 20090461293, by special grade of the financial support from China Postdoctoral Science Foundation Grant No. 201003670, by the open foundation for the research center of resource environment economics in China University of Geosciences (Wuhan) and by the open foundation for Key Laboratory of Tectonics and Petroleum Resources (China University of Geosciences), Ministry of Education No. TPR-2011-11.